\documentclass[conference]{IEEEtran}
\IEEEoverridecommandlockouts
\usepackage{cite}
\usepackage{amsmath,amssymb,amsfonts}
\usepackage{algorithmic}
\usepackage{textcomp}
\usepackage{xcolor}
\usepackage{array}%

\usepackage{times}
\usepackage{soul}
\usepackage{url}
\usepackage[small]{caption}
\usepackage{graphicx}
\usepackage{amsmath}
\usepackage{booktabs}
\newtheorem{example}{Example}[section]

\usepackage[linesnumbered,ruled,vlined]{algorithm2e}
\usepackage{amssymb}
\usepackage{multirow}
\usepackage{color}
\usepackage{comment}

\def\BibTeX{{\rm B\kern-.05em{\sc i\kern-.025em b}\kern-.08em
    T\kern-.1667em\lower.7ex\hbox{E}\kern-.125emX}}
\begin{document}

\title{Learning to Sample: an Active Learning Framework\\
}

\author{\IEEEauthorblockN{Jingyu Shao, Qing Wang and Fangbing Liu}
\IEEEauthorblockA{\textit{Research School of Computer Science} \\
\textit{Australian National University}\\
Acton, ACT, Australia \\
\{jingyu.shao, qing.wang, fangbing.liu\}@anu.edu.au}

}

\maketitle

\begin{abstract}

Meta-learning algorithms for active learning are emerging as a promising paradigm for learning the ``best'' active learning strategy. However, current learning-based active learning approaches still require sufficient training data so as to generalize meta-learning models for active learning. This is contrary to the nature of active learning which typically starts with a small number of labeled samples. The unavailability of large amounts of labeled samples for training meta-learning models would inevitably lead to poor performance (e.g., instabilities and overfitting).
In our paper, we tackle these issues by proposing a novel learning-based active learning framework, called \emph{Learning To Sample} (LTS). This framework has two key components: a sampling model and a boosting model, which can mutually learn from each other in iterations to improve the performance of each other. Within this framework, the sampling model incorporates uncertainty sampling and diversity sampling into a unified process for optimization, enabling us to actively select the most representative and informative samples based on an optimized integration of uncertainty and diversity. To evaluate the effectiveness of the LTS framework, we have conducted extensive experiments on three different classification tasks: image classification, salary level prediction, and entity resolution. The experimental results show that our LTS framework significantly outperforms all the baselines when the label budget is limited, especially for datasets with highly imbalanced classes. In addition to this, our LTS framework can effectively tackle the cold start problem occurring in many existing active learning approaches.

\end{abstract}
\medskip
\begin{IEEEkeywords}
active learning, meta-learning, uncertainty sampling, diversity sampling, boosting
\end{IEEEkeywords}

\medskip
\section{Introduction}
\label{sec:Intro}

Sampling is a fundamental technique for acquiring training data in machine learning applications. However, obtaining large amounts of manually labeled samples is often expensive or simply infeasible in practice. To alleviate this issue, active learning has been extensively studied in the past decades \cite{settles2010active}, which aims to select fewer labeled samples to train a machine learning model as effectively as possible, achieving similar or greater accuracy. At its core, active learning seeks for the most representative or informative samples to be labeled for training by leveraging observations from previously labeled samples \cite{donmez2008paired, maystre2017just, deng2018adversarial}. 

To date, various active learning techniques have been developed from different perspectives \cite{settles2010active}, such as uncertainty sampling \cite{yang2015multi,tong2001support}, query-by-committee \cite{seung1992query}, error or variance minimization \cite{roy2001toward, hoi2006batch}, and expected model change \cite{cai2017active}. They all attempted to address a key challenge in active learning: given a dataset, how to decide which samples in the dataset are more representative or informative than the others for training a machine learning model? However, as evidenced by the experiments presented in these works, there is no one-fit-all solution for active learning. Due to the variety of datasets and machine learning models, different active learning techniques may perform best in different circumstances, depending on the dataset at hand and the machine learning model being chosen.

\begin{figure}[t!]
	\centering
		\includegraphics[height=0.3\textheight]{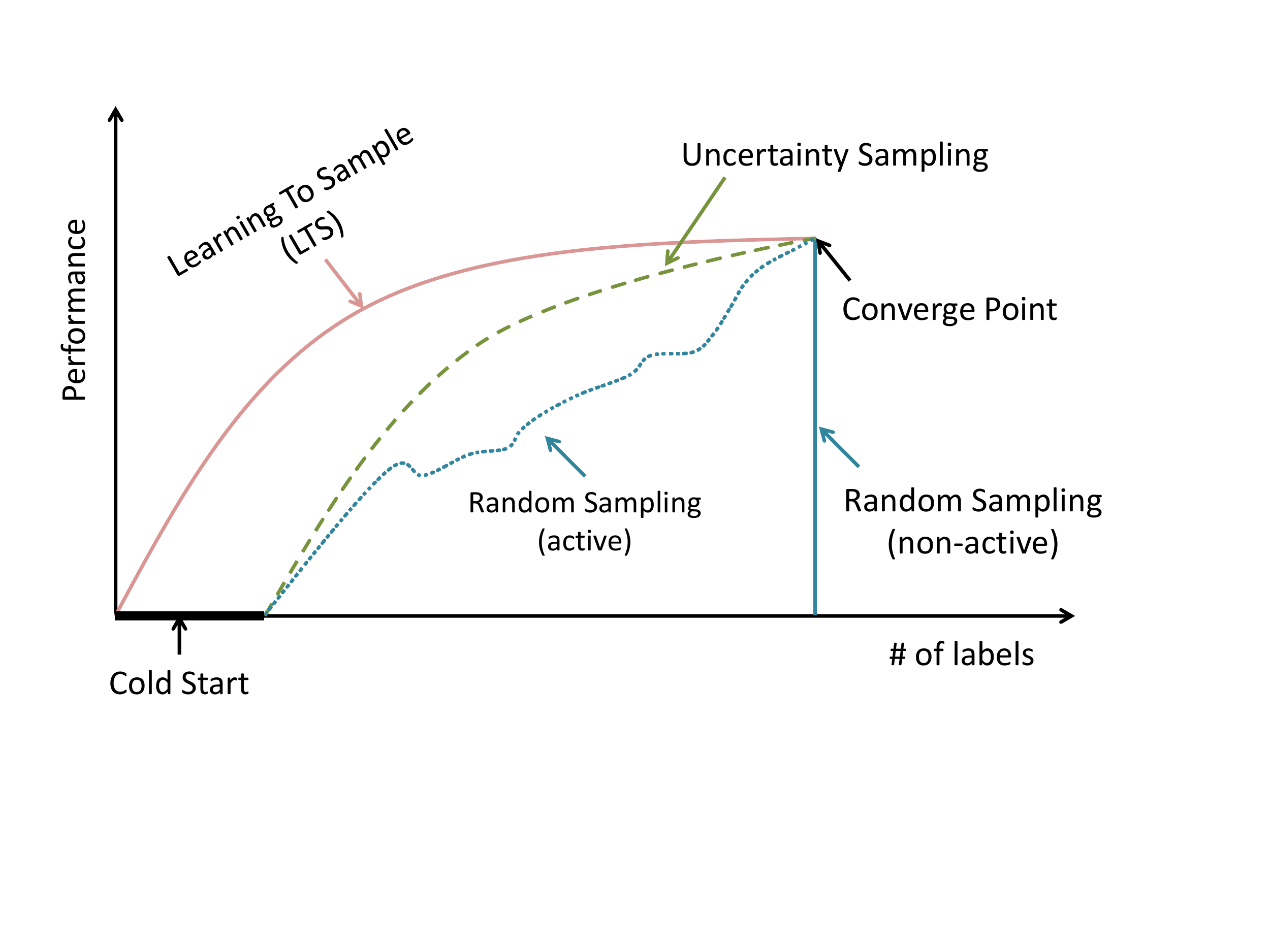}
		\vspace{-20mm}
\caption{An illustration of Learning To Sample (LTS) in relation to uncertainty sampling and random sampling, where random sampling (active) indicates that random samples are gradually selected during the iterations of active learning, and random sampling (non-active) indicates that all samples are randomly selected in a one-off manner (i.e., no active learning).} 
	\label{fig_illustration}	
\end{figure}

Recently, several learning-based active learning approaches have been proposed to address such limitations \cite{hsu2015active, konyushkova2017learning}. Instead of using pre-defined strategies for active learning, these works considered to learn the ``best'' active learning strategy based on the estimated model performance of a meta-learning model. For example, Hsu and Lin \cite{hsu2015active} developed an approach to learn from the performance of a set of active learning strategies adaptively so as to decide a desired active learning strategy. Konyushkova et al. \cite{konyushkova2017learning} proposed a learning based approach using the Monte Carlo method to predict the reduction of generalization error by each unlabeled instance.
Nevertheless, these learning-based active learning approaches still require sufficient training data so as to generalize a meta-learning model. On the contrary, active learning typically starts with a small number of labeled samples (i.e., seed samples) and gradually adds more labeled samples through an iterative learning process. Thus, a meta-learning model can only be trained on a small number of labeled samples at the beginning, which leads to poor performance (e.g., instabilities and overfitting).

In this paper, we aim to propose a learning-based active learning framework to enable a unified sampling process for selecting representative and information samples from different perspectives. Different from the previous active learning approaches, we ground our work based on the following observations: (1) Although uncertainty sampling is one of the widely used active learning techniques \cite{lewis1994sequential}, uncertainty sampling alone tends to select samples that are similar to each other, i.e., samples being selected from a sample space often have similar features \cite{yang2015multi}. (2) Diversity sampling targets to select samples of different kinds (e.g., samples with different features), which is complementary to uncertainty sampling. Thus, the obstacle of uncertainty sampling can be circumvented by combining uncertainty sampling and diversity sampling into a unified sampling process. (3) To find the ``best'' way to integrate these two sampling strategies, meta-learning is a powerful tool, which can optimize this integration process by learning hints from the chosen machine learn models and datasets.

Based on the above observations, we design a novel learning-based active learning framework, called \emph{Learning To Sample} (LTS). In a nutshell, the LTS framework consists of two key components: a sampling model $G$ and a boosting model $F$, which are learned iteratively, and their results can mutually strength each other in iterations. As illustrated in Fig.~\ref{fig_illustration}, the goal of this LTS framework is to help machine learning models achieve better performance with less training data by providing a learning-based active learning process. The design of the LTS framework incorporates the uncertainty and diversity aspects of sampling into a unified process, which can also circumvent the cold start problem \cite{deng2018adversarial,konyushkova2017learning}.   





\medskip
\noindent\textbf{Contributions}
In summary, the contributions in this work are as follows:

\begin{itemize}

        \item We propose a novel active learning framework, namely \emph{Learning To Sample} (LTS), in which a boosting model $F$ and a sampling model $G$ can dynamically learn from each other in iterations for improving the performance of each other.
    
    \item Our sampling model incorporates uncertainty and diversity of samples into a unified process for optimization. This allows us to actively select samples based on the joint impacts of probabilities of being mis-classified by a boosting model and the distribution of samples in a sample space. 

    \item The experimental results show that our active learning approach significantly outperforms all the baselines when the label budget is limited, especially for those datasets with highly imbalanced classes. It also shows that our approach can effectively tackle the cold start problem. 
    
    
\end{itemize} 

\smallskip
It is worth noting that, technically, the boosting model $F$ can be replaced by any classification model and the regressors in the sampling model $G$ can be replaced by any regression model. Thus, the LTS framework is indeed not restricted to specific machine learning models used for classification and regression. 

\section{Learning To Sample Framework}

In this section, we present our learning based active learning framework, called \emph{Learning To Sample} (LTS).


\begin{figure}[t!]
	\centering
		\includegraphics[height=0.35\textheight]{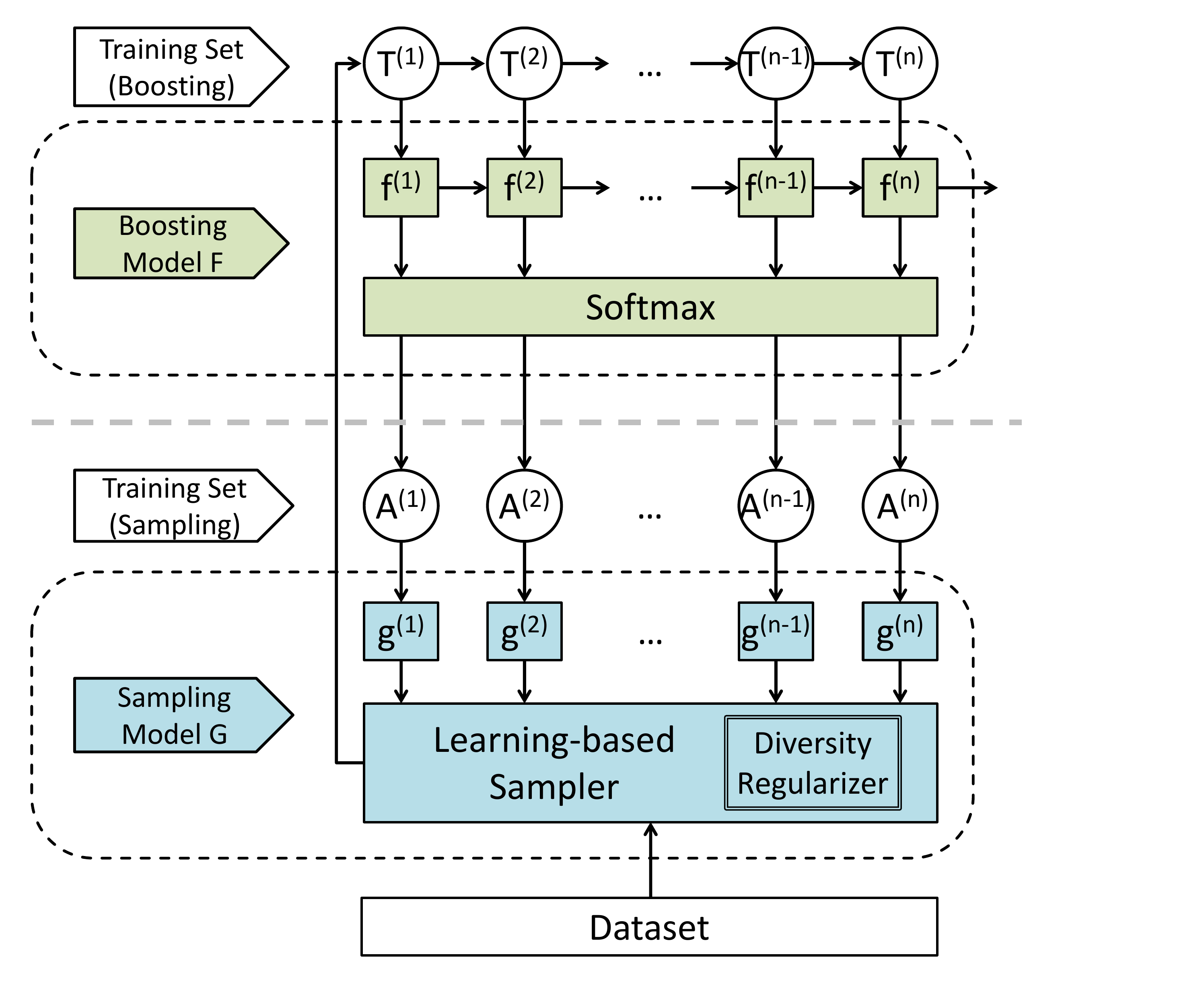}
\caption{The overall framework of Learning To Sample (LTS)}
	\label{fig_frame}	
\end{figure}

As illustrated in Figure \ref{fig_frame}, the LTS framework has two key components: a boosting model $F$ (highlighted in green) and a sampling model $G$ (highlighted in blue). Accordingly, there are two learning processes that are closely coupled: 
(1) learning the boosting model $F$, and (2) learning the sampling model $G$. Specifically, a boosting model $F$ aims to create a strong learner based on a set of weak learners. Thus, the boosting model $F$ is trained iteratively on a sequence of incrementally built training sets in order to add new functions for improving its model performance. Samples in these training sets are actively selected by the sampling model $G$ which is dynamically learned from the performance of the boosting model $F$ during its iterative training process. In the following, we discuss the boosting model and the sampling model in detail.

\subsection{Boosting Model}\label{sec:boosting}
\label{sec:PD}

Let $X\subseteq \mathbb{R}^d$ be a dataset with $|X|$ instances and $\zeta$ be a budget on the total number of instances from $X$ that can be labeled by a human oracle. A training set $T = \{(x_i, y_i)\}^{|T|}_{i=1}$, where $x_i\in X$ and $y_i\in \mathbb{R}$, consists of a set of instances from $X$ and their labels from $\mathbb{R}$. This training set $T$ is incrementally built as the boosting model interacts with the sampling model, i.e., a sequence of training subsets $\langle T^{(1)}, \dots, T^{(n)}\rangle$ such that $T^{(1)} \subseteq T^{(2)} \subseteq \dots \subseteq T^{(n)}$, $T^{(n)}=T$, and $|T^{(n)}|\leq \zeta$, where $T^{(t)}$ for $t\in [1,n]$ is a training subset being used for training the boosting model at the $t$-th iteration.

A boosting model $F$ trains a sequence of functions $\langle f^{(1)}, \dots, f^{(n)}\rangle$ in an additive manner, where $f^{(t)}$ for $t\in [1,n]$ is a function being added into $F$ at the $t$-th iteration. More specifically, the individual results of the first $t$-$1$ functions are combined to predict the label of an instance at the $(t$-$1)$-th iteration such that:
\begin{equation}
  \hat{y}_{i}^{(t-1)} = \sum_{k=1}^{t-1} f^{(k)} (x_i).
\end{equation}
Then, the $t$-th function $f^{(t)}$ is trained on the actively selected training subset $T^{(t)}$ by minimizing the following objective function:

\begin{equation}
\label{eq_subobj}
\begin{aligned}
& \sum_{(x_{i}, y_{i}) \in T^{(t)}}  \ell_1(\hat{y}^{(t-1)}_{i}+f^{(t)}(x_i), {y_{i}}) + \Omega_1(f^{(t)})\\
\end{aligned}
\end{equation}
where $\ell_1$ is a differentiable loss function and $\Omega_1(f^{(t)})$ is the penalty for the complexity of $f^{(t)}$.

After the $t$-th function $f^{(t)}$ is learned, the boosting model $F$ sends its feedback to the sampling model $G$ via a softmax layer. This allows the sampling model $G$ to leverage hints from the prediction results of $\langle f^{(1)}, \dots,f^{(t)}\rangle$ and actively select the most informative instances as new samples for the next iteration, leading to $T^{(t+1)}$. We use the \emph{Softmax} function \cite{sutton1998reinforcement} to obtain probabilities of being mis-classified for training samples. Specifically, in the $t$-th iteration, the softmax layer takes $\mathbf{l}^{(t)}=\langle \ell (\hat{y}^{(t)}_{1}, {y_{1}}), \dots, \ell (\hat{y}^{(t)}_{q}, {y_{q}})\rangle$ as input, where $q=|T^{(t)}|$ and each $\ell (\hat{y}^{(t)}_{j}, {y_{j}})$ in $\mathbf{l}^{(t)}$ refers to the loss of a training sample $x_j$ from $T^{(t)}$, then generates $\mathbf{z}^{(t)}=\langle z^{(t)}_1, \dots, z^{(t)}_q\rangle$, i.e.,
\begin{equation}\label{equ:softmax}
{z}^{(t)}_i=\text{\emph{Softmax}}({l}_i^{(t)}),
\end{equation}
where $\emph{Softmax}(l_{i}^{(t)})=e^{l_{i}^{(t)}}/\sum_{j=1}^q e^{l_{j}^{(t)}}$ and $l_i^{(t)}=\ell (\hat{y}^{(t)}_{i}, {y_{i}})$.

\vspace*{0.3cm}
\subsection{Sampling Model}\label{sec:sampling}

Let $X^{(t)}_L=\{x_i\in X|(x_i,y_i)\in T^{(t)}\}$ be the set of labeled instances and $X^{(t)}_U=X-X^{(t)}_L$ be the set of unlabeled instances in the $t$-th iteration. A sampling model $G$ aims to select a set $\Delta^{(t)}$ of the most informative samples from unlabeled instances at the $t$-th iteration such that $X^{(t+1)}_L=X^{(t)}_L\cup \Delta^{(t)}$ and $X^{(t+1)}_U=X^{(t)}_U- \Delta^{(t)}$. Consequently,  $T^{(t+1)}=T^{(t)}\cup\{(x_i,y_i)|x_i\in \Delta^{(t)}\}$ is generated and sent to the boosting model $F$ for training the function $f^{t+1}$. 

The question arising here is: how to actively select a set $\Delta^{(t)}$ of the most informative samples at the $t$-th iteration? In the LTS framework, two kinds of samples are primarily targeted: (1) samples that are likely to be mis-classified by the boosting model; (2) samples that have diverse features in the sample space. They relate to the uncertainty and diversity aspects of sampling, respectively. Hence, at the $t$-th iteration, the sampling model $G$ learns to select a set $\Delta^{(t)}$ of most informative samples by maximizing the following objective:

\begin{equation}
\label{eq_g}
\begin{aligned}
& \underset{}{\text{\textbf{maximize}}}
& & \sum^{k}_{i=1} v_{i} g^{(t)}(x_{i}) + \alpha \times \Gamma(\mathbf{v}) \\
& \text{\textbf{subject to}}		
& &  ||\mathbf{v}||_1 =|\Delta^{(t)}|  \\	
&&& 	
\end{aligned}
\end{equation}
where $k=|X^{(t)}_U|$, $\mathbf{v} = (v_1, ..., v_k)^T \in \{0, 1\}^k$, and each $v_i$ is associated with an instance $x_i\in X^{(t)}_U$. When $v_i = 1$, it indicates that $x_i$ is selected as a sample, and conversely, $v_i=0$ indicates that $x_i$ is not selected. The term $g^{(t)}(x_{i})$ indicates the uncertainty score of an instance $x_i$ which is predicated by a regressor $g^{(t)}$, and the regularization term $\Gamma(\mathbf{v})$ controls the distribution of selected instances in order to ensure their diversity in the sample space.
$\alpha$ is a parameter used for balancing the impacts of uncertainty and diversity on samples, i.e., $\alpha >1$ indicates that diverse samples are preferred, while $\alpha < 1$ indicates that samples with high probabilities of being mis-classified are preferred.
Further details for our sampling model will be discussed in the next section.

\medskip

\section{Sampling Strategies}

In the following, we discuss how the sampling model $G$ handles the uncertainty and diversity aspects of samples.
We first present an uncertainty sampling strategy by training a regressor $g^{(t)}$ in each iteration, then describe how the regularization term $\Gamma(\mathbf{v})$ is used to deal with diversity sampling.

\begin{figure} 
		\centering
	\begin{center}
		\includegraphics[height = 0.27\textheight]{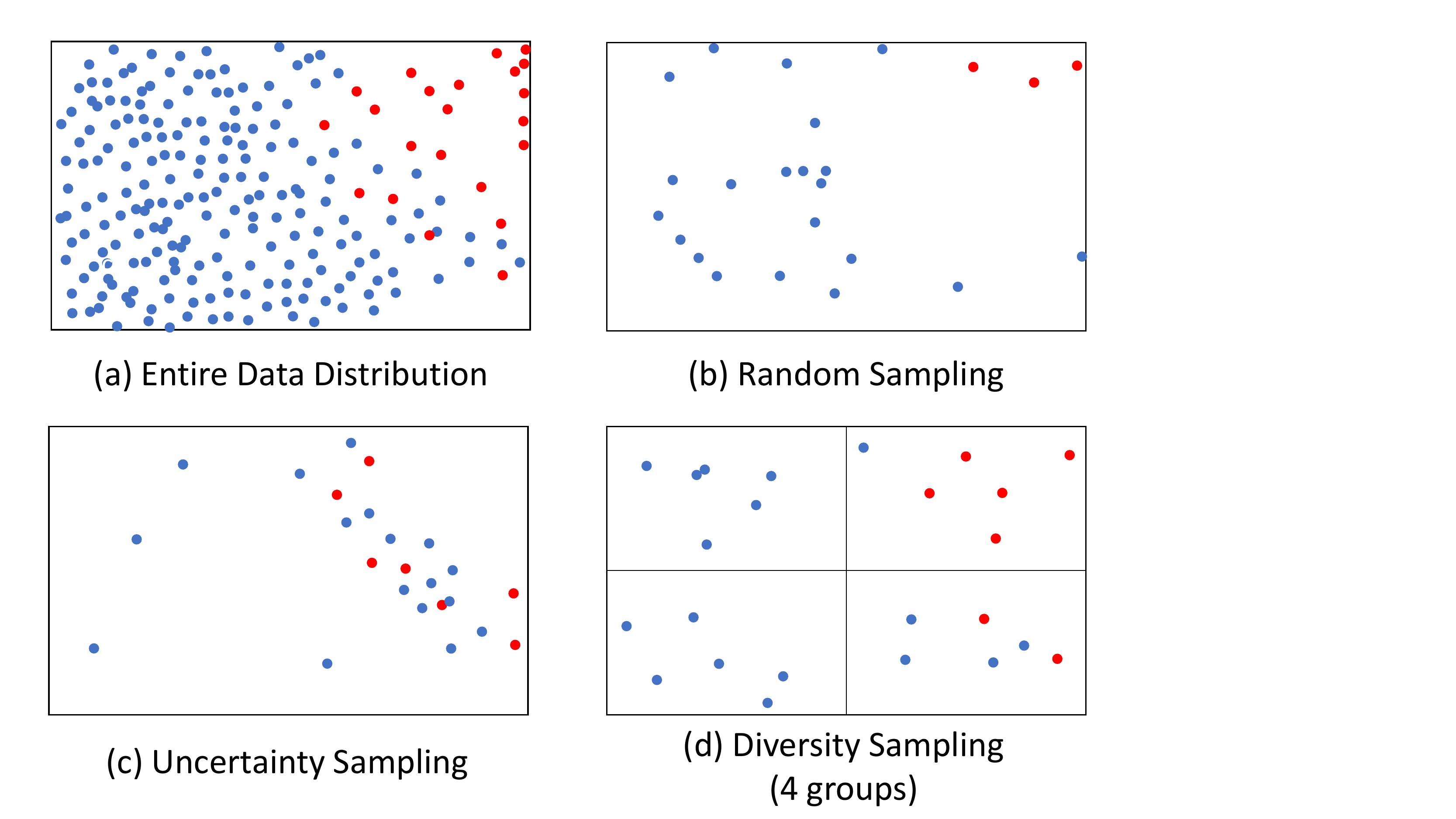}
	\end{center}
	\caption{Comparison of different sampling strategies, where 24 samples are selected in each of (b), (c) and (d).} 
	\label{fig_sampling}
\end{figure}

Figure~\ref{fig_sampling} illustrates our sampling strategies, i.e. uncertainty sampling and diversity sampling, in comparison with random sampling. Figure \ref{fig_sampling}.(a) describes a real data distribution with two classes (red and blue). Figure \ref{fig_sampling}.(b) shows that random sampling can only select very few samples from the minority class (red). Figure \ref{fig_sampling}.(c) shows using uncertainty sampling leads to samples that are similar. Figure \ref{fig_sampling}.(d) shows that diversity sampling can evenly select samples from different groups in the sample space.

\subsection{Uncertainty Sampling} 
In the LTS framework, we predict the uncertainty of instances by learning from the performance of the boosting model, i.e. the training loss. 
We dynamically construct a training dataset to train a regressor for predicting the uncertainty in each iteration. 

Formally, a training set $A^{(t)}$ for the sampling model $G$ is constructed at the $t$-th iteration such that $A^{(t)}=\{(x_i, z_i^{(t)})|(x_i,y_i) \in T^{(t)}, z^{(t)}_i \in [0,1]\}$, where $\mathbf{z}^{(t)}=\langle z^{(t)}_1,\dots, z^{(t)}_{q}\rangle$ is generated by the softmax layer of the boosting model $F$ and $q=|T^{(t)}|$ as shown in Eq.~\ref{equ:softmax}. 
Thus, each training set $A^{(t)}$ contains the same set of instances as in $T^{(t)}$, but the labels of these instances in $A^{(t)}$ are different from the labels in $T^{(t)}$. 
Furthermore, each label $z_i^{(t)}$ represents the probability of being mis-classified of an instance $x_i$ after the first $t$ iterations. 
We then predict the uncertainty score $g^{(t)}(x_i)$ of an unlabeled instance $x_i\in X_U^{(t)}$ in Eq. \ref{eq_g} by solving a regression problem, i.e., training $g^{(t)}$ to minimize the following objective in the $t$-th iteration:
\begin{equation}
	\label{eq_transg}
	\sum_{(x_i, z^{(t)}_{i})  \in A^{(t)}} w^{(t)}_{i}\ell_2 (g^{(t)}(x_i), z^{(t)}_i) + \Omega_2(g^{(t)})
\end{equation}
where $\ell_2$ is also a differentiable loss function, $\Omega_2(g^{(t)})$ is the penalty for the complexity of $g^{(t)}$, and $w^{(t)}_i$ is a weighted value for $x_i$ and is dynamically adjusted during the iterations. The intuition behind $w^{(t)}_i$ is to give higher weighted values to samples that are uncertain in more iterations, rather than samples that are uncertain in fewer iterations. 
For example, if a sample is mis-classified by the boosting model for a number of times, it will be assigned a higher weighted value than another sample which is mis-classified only once.
We will present a method of assigning dynamic weighted values in Section \ref{sec:details}.

\subsection{Diversity Sampling} 


 In the LTS framework, we deal with the diversity of samples by partitioning the sample space into a number of different groups such that instances in the same group are more similar than the instances in different groups. Then we use the regularization term $\Gamma(\mathbf{v})$ in Eq. \ref{eq_g} to regulate the sampling model, i.e., selecting samples from each group evenly. 

Suppose that unlabeled instances in $X^{(t)}$ are partitioned into a set of groups $\{X^{(t)}_{1}, \dots, X^{(t)}_{b}\}$ alike in certain features. 
Then we define the regularization term $\Gamma(\mathbf{v})$ over $\{X^{(t)}_{1}, \dots, X^{(t)}_{b}\}$ using a $l_{2,1}$-norm function as:
\begin{equation}
  \Gamma(\mathbf{v}) = ||\mathbf{v}||_{2,1} = \sum^{b}_{j = 1} ||\mathbf{v}_{j}||_{2}
\end{equation}
where $b$ is the total number of groups associated with $X^{(t)}_U$, $\mathbf{v}$ is partitioned into $\{\mathbf{v}_{1}, \dots, \mathbf{v}_{b}\}$ where $\Sigma_{j=1}^b |\mathbf{v}_j|=|\mathbf{v}|$, $\mathbf{v}_j\in \{0,1\}^m$, $m=|X^{(t)}_{j}|$ and $j\in [1,b]$. That is, $||\mathbf{v}_{j}||_{2}$ is the $l_2$-norm of $\mathbf{v}_j$ that is a binary vector whose elements correspond to instances in group $X^{(t)}_j$.



It is known that the $l_{2,1}$-norm favors on selecting samples with diversity \cite{jiang2014self}. When the value of the $l_{2,1}$-norm is small, non-zero entries of $\mathbf{v}$ are concentrated in a small number of groups, i.e. the distribution of samples is limited to a small number of groups and accordingly the diversity of samples is low. On the contrary, when maximizing the $l_{2,1}$-norm in Eq.~\ref{eq_g}, there is a counter-effect on the distribution of samples, i.e. non-zero entries of $\mathbf{v}$ are widely distributed w.r.t. as many groups as possible and thus the diversity of samples is high.


\smallskip
\begin{example}
  Consider Figure~\ref{fig_sampling}(d) in which the sample space is partitioned into four groups and a number of 24 samples will be selected. If we select 6 samples from each group, $||\mathbf{v_j}||_2=\sqrt{6}$, we have $\Gamma(\mathbf{v}) = \sum^{4}_{j = 1} ||\mathbf{v}_{j}||_{2} = \sqrt{6} \times 4 = 9.8$. If we select 24 samples from only one group, $||\mathbf{v_j}||_2=\sqrt{24}$, then $\Gamma(\mathbf{v}) = \sum^{1}_{j = 1} ||\mathbf{v}_{j}||_{2} = \sqrt{24} = 4.9$.
\end{example}
\smallskip

\section{Algorithm Description}
\label{sec:details}
In this section, we propose an algorithm for the LTS framework and discuss several important aspects of this algorithm which may influence the effectiveness of sampling. 


\begin{algorithm*}
	\KwIn{\hspace{2.5 mm} $X$ with $k$ groups, i.e. $\sum^k_{i=1}X_i^{(0)}=X$; label budget $\zeta$;\\
		\hspace{13.5 mm} Balancing parameter $\alpha$; Number of iterations $n$;\\
	}
	\KwOut{
		A boosting model $F$
	}	
    Initialize $T^{(0)} = \emptyset$\\ 
    Select a set of seed samples $\Delta^{(0)}$ from $k$ groups to maximize $\Gamma(\textbf{v})$ , where $|\Delta^{(0)}| = \frac{\zeta}{n}$ \\
    \For {t = 1, \dots, n}{
    Update $T^{(t)} = T^{(t-1)} +　\Delta^{(t-1)}$\\ 
      Train an additive function $f^{(t)}$ by minimizing the objective in Eq.~\ref{eq_subobj} using $T^{(t)}$ \\
      Generate a training set $A^{(t)}$ \\
      Train a regression function $g^{(t)}$ by minimizing the objective in Eq.~\ref{eq_transg} using $A^{(t)}$\\
      Update $X_{i}^{(t)} =\{x\in X_{i}^{(t-1)}| x\notin\Delta^{(t-1)}\}$, where $i=1,\dots, k$ \\
      Select a set of samples $\Delta^{(t)}$ from $\sum^k_{i=1}X_{i}^{(t)}$ by maximizing the objective in Eq.~\ref{eq_g}, with $|\Delta^{(t)}| = \frac{\zeta}{n}$

    }
	\caption{Learning To Sample (LTS)}
	\label{Algo:LS}	
\end{algorithm*}

A high-level description of the algorithm is presented in Algorithm~\ref{Algo:LS}. This algorithm takes a k-grouped dataset, a label budget and the number of iterations as input. The first step is to initialize the training set $T^0$ and select a set of seed samples from $k$ groups using our diversity sampling strategy (Lines 1-2). Then the algorithm iterates to train a boosting model by actively selecting samples (Lines 4-9). For each $t$-th iteration, we first update the training set $T^{(t)}$ by adding newly selected samples $\Delta^{(t-1)}$ into the previous training set $T^{(t-1)}$ (Line 4). Then an additive function $f^{(t)}$ is trained for the boosting model $F$ (Line 5). After that, a new training set $A^{(t)}$ is generated for the sampling model $G$ based on the output of the current $F$ (Line 6), and a regressor is trained for uncertainty prediction (Line 7). We then update the groups $\{X_1^{(t)}, \dots, X_k^{(t)}$ by excluding the previous selected samples in $\Delta^{(t-1)}$, and select a new set of samples $\Delta^(t)$ based on Eq.~\ref{eq_g} Eq.~\ref{eq_g} (Lines 8 - 9). The algorithm finally yields a trained boosting model as output.

In the following, we first focus on discussing three important aspects of the algorithm: (i) How to decide dynamic weighted values for samples? (ii) How to partition a sample space into different groups? (iii) How to distribute a given label budget across iterations? Then, we will discuss how the cold start problem can be alleviated by our algorithm. 

\subsection{How to decide dynamic weighted values for samples?} 
During the training process of the boosting model, some samples in the training set may have high training losses in a number of iterations. Such samples are often informative for predicting uncertainty. Thus, a dynamic weighted value $w^{(t)}_i$ is assigned to each sample $x_i$ to indicate its importance, as shown in Eq. \ref{eq_transg}. By extending the work by Freund and Schapire \cite{freund1997decision}, we develop the following method of assigning dynamic weighted values in the LTS framework. In each iteration, dynamic weighted values of samples are updated in two steps: 
\begin{itemize}
\item[(1)] \textbf{Initialization: } For each new sample $x_i$ at the $t$-th iteration, i.e. a sample in $\Delta^{(t-1)}$, we have: 
\begin{equation}
w_{i}^{(t-1)} = \frac{1}{|\Delta^{(t-1)}|}.
\end{equation}

\item[(2)] \textbf{Adjustment: } Then, the weighted value for each sample $x_i$ in $A^{(t)}$ is re-calculated as: 
\begin{equation}
\label{eq_wt}
w_{i}^{(t)} = w_{i}^{(t-1)} \times \frac{e^{-\frac{1}{2}ln(\frac{1-\epsilon^{(t-1)}}{\epsilon^{(t-1)}}){g^{(t-1)}(x_i) z^{(t-1)}_{i} }}}{Z_t},
\end{equation}

where $\epsilon^{(t-1)} = \frac{\sum_i{z^{(t-1)}_{i}}}{|T^{(t-1)}|}$ and $Z_t$ is a normalization factor ensuring that the sum of all weighted values of samples in $A^{(t)}$ equals to $1$.
\end{itemize}

In our algorithm, a regressor $g^{(t)}$ is iteratively trained by minimizing the objective in Eq.~\ref{eq_transg}, in which dynamic weighted values are updated using the above method in each iteration.

\smallskip
\subsection{How to partition a sample space into groups?}
A key challenge of diversity sampling is: how to partition a sample space into groups such that instances in the same group are more similar than instances in different groups? In many real-world applications, samples that have same features are likely to be more similar than samples that have different features. Thus, we consider to partition a sample space based on available features of samples. This can also avoid common issues of sampling based on a data distribution, such as selecting too many similar samples from high density areas. In doing so, diversity sampling in our algorithm can select samples that are complementary to ones being selected by uncertainty sampling.  

Formally, given a sample space with $d$ features, a label budget $\zeta$ and a number $n$ of iterations, we partition the sample space into $k$ groups where $k = {\lceil \sqrt[d]{\frac{\zeta}{n}} \rceil}^d $ and $\lceil \hspace{0.1cm} \rceil$ indicates the ceiling function. For example, if we have $\zeta=600$, $n = 20$ and $d=4$, then $k = {\lceil \sqrt[4]{\frac{600}{20}} \rceil } ^4 = {\lceil 2.34 \rceil } ^4 = 81$, i.e., 81 groups. Each of such groups corresponds to an area in the sample space and samples from the same area have some common features.  


\smallskip
\subsection{How to distribute label budget across iterations?}
Under a given label budget $\zeta$, when more samples are selected at the beginning of the training process, it implies that less samples can be used in the later iterations to leverage hints from observed samples for improving performance. For example, when $|\Delta^{(1)}| = \zeta$, i.e., all samples are used in the first iteration, the training process in the LTS framework would be the same as in the traditional training process. On the other hand, if allocating more samples to the later iterations, the boosting model $F$ would have higher variance in the early iterations, but a better chance to ''bias'' samples for active learning in the later iterations. 

In our algorithm, we distribute a label budget equally over all iterations, i.e., $|\Delta^{(t)}| = \zeta/n$ for any $t\in[1,n]$ (Line 2 of Algorithm \ref{Algo:LS}). An alternative is to distribute samples in an exponentially decreasing manner over iterations, i.e., $|\Delta^{(t)}| = \zeta/2^t$. As will be discussed in our experiments later, the former approach outperforms the latter one in almost all cases.


\begin{table*}[!htbp]
\caption{Characteristics of datasets}
	\centering	\label{tab:dataset2}
	\begin{tabular}{|c|lccccc|}
		\hline
		Classification Tasks & Datasets & $\#$ Attributes & $\#$ Instances ($|X|$) & $\#$ Classes& Types of Labels & Class Imbalance Ratio  \\
		\hline\hline
		Image classification & Mnist & $28 \times 28$ & 60,000 & 10&10 digits (i.e. 0-9)& N/A \\
		\hline
		Salary level prediction & Adult & 14 & 48,842 &2&\{above 50k, not above 50k\}& 1 : 3 \\
		\hline
		\multirow{4}{*}{Entity resolution}&Cora & 12 & 837,865 &2& \{match, non-match\} &1 : 49 \\
		&DBLP-Scholar & 4 &  168,112,008 & 2& \{match, non-match\} & 1 : 71,233 \\
		&DBLP-ACM & 4 &  6,001,104 &2& \{match, non-match\} & 1 : 2,698 \\		
		&NCVoter & 18 &  10M & 2&\{match, non-match\} & 1:420 \\
		\hline
	\end{tabular}
		\vspace{-5mm}
\end{table*}

\subsection{Discussion}

As reported in the previous works \cite{deng2018adversarial,konyushkova2017learning}, the \emph{cold start} problem often occurs in active learning because only a small amount of labeled samples is available in early iterations. Essentially, this is due to the inability of making reliable predictions by a machine learning model if training data is not sufficient. When a dataset has highly imbalanced classes (i.e., the number of instances from a majority class is much more than the number of instances from a minority class), the cold start problem can be further aggravated. Treating samples of all classes equally often leads to selecting samples that are likely to be similar or highly correlated, and thus are not representative \cite{jiang2014self, yang2015multi}.
 
In the LTS framework, the uncertainty of samples is measured using a regressor that is dynamically trained on samples labeled with their losses from the boosting model. If we select samples by only taking the uncertainty of samples into consideration, the cold start problem would also occur in our work. Since one of the reasons underlying the cold start problem is that training data is too small to be representative, we thus partition a sample space into a number of groups based on similarity of features and introduce the regularization term $\Gamma(\mathbf{v})$ to ensure that more representative samples are selected from such a k-grouped sample space. Our experiments show that this approach works effectively for addressing the cold start problem (the experimental results will be discussed later in Section \ref{sec:experiments}).




\section{Experiments}\label{sec:experiments}
 We have conducted experiments to empirically verify our LTS approach, aiming to answer the following questions: 
 \begin{itemize}
     \item[(1)] Given a limited label budget, how does our LTS approach perform in comparison with other sampling methods? 
     \item[(2)] How effectively can our LTS approach deal with the cold start problem and the class imbalance problem? 
     \item[(3)] How does the balancing parameter $\alpha$ affect the performance of our LTS approach? 
     \item[(4)] How do two sampling distribution methods perform, i.e. equal distribution vs exponentially decreasing distribution? 
     \item[(5)] How does our LTS approach perform in reducing label budgets while still achieving the same level of quality for classification as other sampling methods?
 \end{itemize}

\subsection{Experimental Setup}
We evaluate our LTS framework on three different classification tasks: image classification, salary level prediction, and entity resolution \cite{shao2018active}. The first is a multi-class classification task, while the other two are binary classification tasks.

\medskip
\noindent\textbf{Datasets. } Six datasets are used in our experiments: 
(1) \emph{Mnist}\footnote[1]{Available from: \emph{http://yann.lecun.com/exdb/mnist/}} dataset contains $28 \times 28$ images, and each image corresponds to a handwritten digit. The task is to classify the images into ten categories, i.e. from 0 to 9. 
(2) \emph{Adult}\footnote[2]{Available from: \emph{https://archive.ics.uci.edu/ml/datasets/adult}} dataset contains adults' personal information. The task is to predict if a person's salary income is more than 50k. 
(3) \emph{Cora}\footnote[3]{Available from: \emph{http://secondstring.sourceforge.net}} 
dataset contains bibliographic records of machine learning publications.  
(4) \emph{DBLP-Scholar}\footnotemark[3] dataset contains bibliographic records from the DBLP and Google Scholar websites.
(5) \emph{DBLP-ACM} \cite{kopcke2010evaluation} dataset contains bibliographic records from the DBLP and ACM websites.
(6) \emph{North Carolina Voter Registration (NCVoter)}\footnote[4]{Available from: \emph{http://alt.ncsbe.gov/data/}} dataset contains real-world voter registration information of people from North Carolina in the USA. 
The datasets (3)-(6) are used for entity resolution, which aims to detect if two records from one or two datasets refer to the same entity (i.e. to classify two records as being a match or a non-match). 

Table~\ref{tab:dataset2} summarizes the characteristics of the above six datasets. We can see that the datasets for entity resolution are highly imbalanced, i.e., the number of instances from the majority class (non-match) is much more than the number of instances from the minority class (match) in these datasets.

\medskip
\noindent\textbf{Baseline methods. }
We use the following baseline methods: 
(1) \emph{CART} \cite{breiman1984classification}, short for Classification And Regression Tree, is a decision tree approach. 
(2) \emph{XG} \cite{chen2016xgboost}, short for eXtreme Gradient Boosting, is a widely used and state-of-the-art boosting approach for decision trees. 
(3) \emph{XG+RS}, refers to applying XG on training sets built using the random sampling strategy. (4) \emph{XG+US}, refers to applying XG on training sets built only using the uncertainty sampling strategy, i.e., $\alpha = 0$ in our LTS framework. (5) \emph{XG+DS}, refers to applying XG on training sets built only using the diversity sampling strategy, i.e., $\alpha \rightarrow \infty$ in our LTS approach. For clarity,
our LTS approach is denoted as \emph{XG+LTS}. To evaluate how the exponentially decreasing distribution of samples may affect performance, we denote a variant of \emph{XG+LTS} as \emph{XG+LTS(E)} which only differs from XG+LTS in distributing samples in an exponentially decreasing manner. By default, we set $\alpha=1$ for XG+LTS and \emph{XG+LTS(E)}, unless otherwise stated.
For XG, the maximum depth of each tree is 5, and other parameters are set as default as used in \cite{chen2016xgboost}.


\medskip
\noindent\textbf{Measures.} We use \emph{accuracy} to evaluate the classification results over the first two datasets, i.e. Mnist and Adult.
As the datasets of entity resolution tasks are highly imbalanced, we use \emph{precision}, \emph{recall} and \emph{f-measure} as measures for entity resolution instead of accuracy. Basically, \emph{recall} is the fraction of true positives among the total number of true matches, \emph{precision} is the fraction of true positives over all positives, and \emph{f-measure} (FM) is the harmonic mean of recall and precision, i.e. $\emph{FM} = \frac{2*Recall*Precision}{Recall + Precision}$. 
\begin{figure}[t!] 
		\centering
	\begin{center}
		\includegraphics[width =0.5\textwidth]{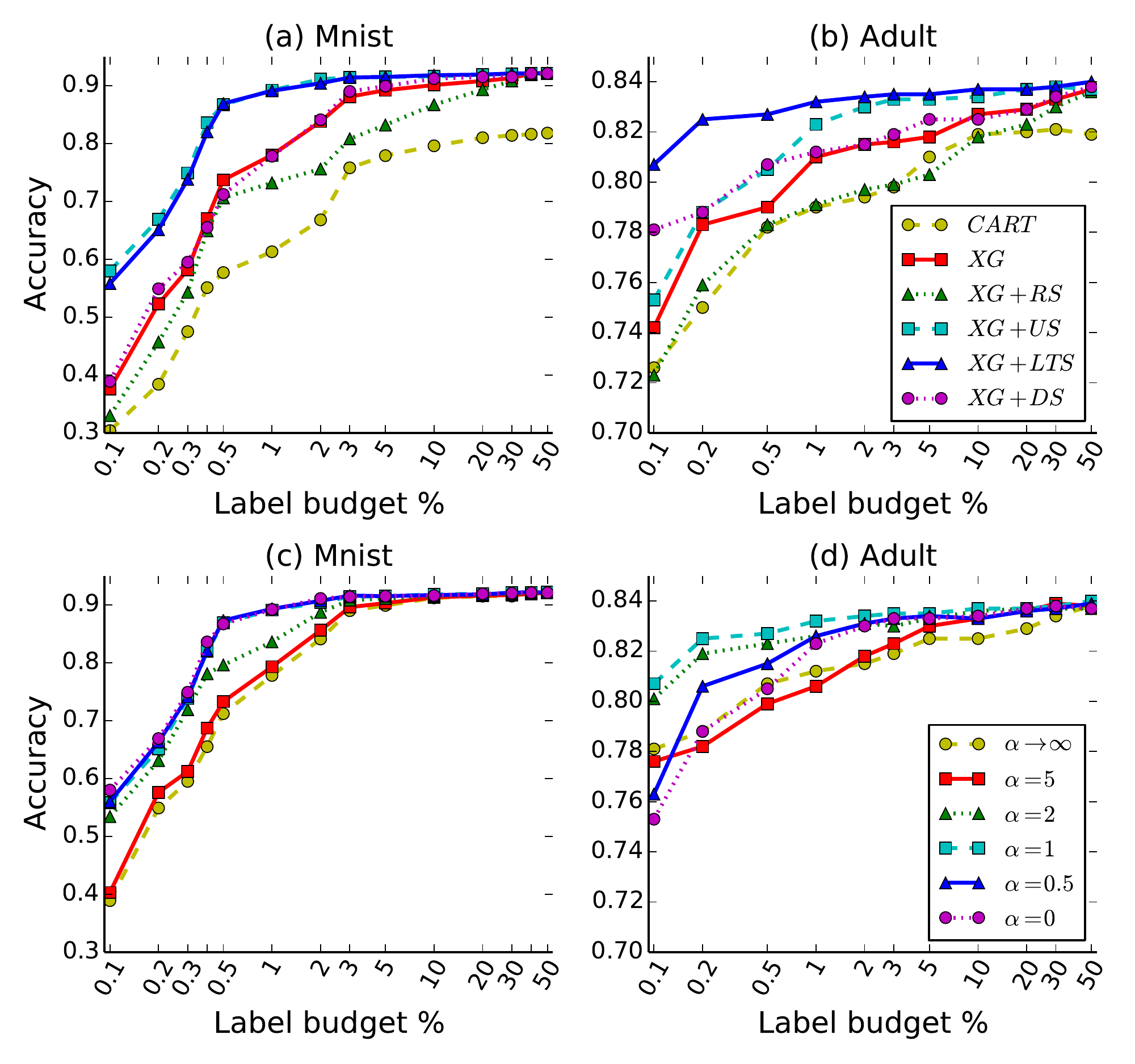}
	\end{center}
	\caption{Comparison of accuracy results for image classification and salary level prediction tasks under different label budgets} 
	\label{fig_accuracy}
\end{figure}

\medskip
\noindent\textbf{Label budgets. }In our experiments, for each dataset $X$, we specify a label budget in terms of a certain percentage of the size of the dataset ($|X|$). For example, when using 1\% as the label budget for the dataset NCVoter, i.e. 1\% of $|X|$, we have 100,000 samples because NCVoter contains 10M instances in total. We also set $n=20$ (i.e., 20 iterations), and distribute a label budget as follows: 
\begin{itemize}
\item For the methods CART and XG, a label budget is used in the first iteration to randomly select all samples within the given label budget for training. 
\item For the methods XG+RS, XG+US, XG+DS and XG+LTS, a given label budget is evenly divided over 20 iterations. For example, given a label budget 1\% for NCVoter, 5,000 samples are used in each iteration for 20 iterations.
\item For the method XG+LTS(E), a given label budget is divided over 20 iterations in an exponentially decreasing manner.
\end{itemize}

\subsection{Results and Discussion}
We discuss our experimental results to answer the aforementioned questions at the beginning of this section.

\smallskip
\subsubsection{Performance under different label budgets}

\begin{table*}[h]

\caption{Comparison of f-measure results for entity resolution tasks under different label budgets}
	\centering
	\label{tab_fm}
	\scalebox{1.0}{
	\begin{tabular}{|c|c||ccc||c|cccc|c||c|}
        \hline
        \multicolumn{1}{|c|}{\multirow{2}{*}{Dataset}} & Label Budget $\zeta$ & \multicolumn{1}{|c|}{\multirow{2}{*}{CART}} & \multicolumn{1}{|c|}{\multirow{2}{*}{XG}} & \multicolumn{1}{|c||}{\multirow{2}{*}{XG+RS}} & XG + US & \multicolumn{4}{|c|}{XG+LTS} & XG + DS & XG + LTS(E) \\  
        \cline{7-10}
        & (\% of $|X|$) & \multicolumn{1}{|c|}{} & \multicolumn{1}{|c|}{} & \multicolumn{1}{|c||}{} & $\alpha = 0$ & \multicolumn{1}{|c|}{$\alpha = 0.5$} & \multicolumn{1}{|c|}{$\alpha = 1$} & \multicolumn{1}{|c|}{$\alpha = 2$} & \multicolumn{1}{|c|}{$\alpha = 5$} & $\alpha \rightarrow \infty$ & $\alpha = 1$\\
        \hline
        \multirow{6}{*}{Cora} &
        0.01 & 0     & 0     & 0     & 0     & 0.637 & 0.857 & 0.861 & 0.867 & \textbf{0.878} & 0.862\\
        &0.05 & 0.741 & 0.763 & 0.750 & 0.827 & 0.851 & 0.864 & 0.870 & 0.883 & \textbf{0.885} & 0.867\\
        &0.1  & 0.788 & 0.796 & 0.787 & 0.823 & 0.863 & 0.862 & 0.873 & \textbf{0.887} & 0.886 & 0.870\\
        &0.5  & 0.848 & 0.835 & 0.835 & 0.873 & 0.893 & \textbf{0.900} & 0.895 & 0.895 & 0.893 & 0.890\\
        &1    & 0.868 & 0.878 & 0.880 & 0.870 & 0.896 & 0.902 & \textbf{0.904} & 0.898 & 0.894 & 0.896\\
        &5    & 0.878 & 0.897 & 0.892 & 0.907 & 0.912 & \textbf{0.915} & 0.913 & 0.902 & 0.898 & 0.904\\
		\hline
        \multirow{6}{*}{NCVoter} &
        0.01 & 0     & 0     & 0     & 0     & 0.403 & 0.324 & 0.403 & 0.752 & \textbf{0.875} & 0.571\\
        &0.05 & 0     & 0     & 0     & 0     & 0.903 & 0.954 & 0.989 & \textbf{0.993} & 0.991 & 0.934\\
        &0.1  & 0     & 0     & 0     & 0     & 0.989 & \textbf{0.994} & 0.993 & 0.993 & 0.993 & 0.993\\
        &0.5  & 0     & 0     & 0     & 0     & 0.993 & \textbf{0.994} & 0.993 & 0.993  & 0.991 & \textbf{0.994}\\
        &1    & 0.334 & 0.379 & 0.398 & 0     & 0.993 & 0.993 & 0.993 & 0.992 & \textbf{0.994} & 0.993\\
        &5    & 0.993 & 0.993 & 0.994 & 0.993 & 0.993 & \textbf{0.997}& 0.993 & 0.994 & 0.993 & 0.994\\
		\hline
        \multirow{6}{*}{} &        
        0.1  & 0     & 0     & 0     & 0     & 0 & 0 & 0 & 0 & \textbf{0.397} & 0\\
        &0.5  & 0     & 0     & 0     & 0     & 0.382 & 0.702 & \textbf{0.720} & 0.651 & 0.632 & 0.679\\
        DBLP-&1    & 0.348 & 0.347 & 0.279 & 0     & 0.813 & \textbf{0.878} & 0.778 & 0.730 & 0.721 & 0.793\\
        ACM&2    & 0.599 & 0.767 & 0.680 & 0.403 & 0.851 & \textbf{0.884} & 0.867 & 0.789 & 0.783 & 0.854\\
        &5    & 0.870 & 0.850 & 0.803 & 0.874 & \textbf{0.935} & 0.931 & 0.889 & 0.837 & 0.833 & 0.891\\
        &10   & 0.903 & 0.911 & 0.890 & 0.926 & \textbf{0.983} & 0.981 & 0.937 & 0.893 & 0.899 & 0.933\\
		\hline
        \multirow{6}{*}{} &        
        0.1  & 0     & 0     & 0     & 0     & 0.586 & 0.723 & 0.733 & \textbf{0.741} & 0.731 & 0.727\\
        &0.5  & 0.378 & 0.54  & 0.498 & 0.555 & 0.764 & 0.773 & \textbf{0.794} & 0.790 & 0.780 & 0.781\\
        DBLP-&1    & 0.562 & 0.669 & 0.659 & 0.738 & 0.793 & 0.804 & \textbf{0.808} & 0.793 & 0.792 & 0.794\\
        Scholar&2    & 0.772 & 0.806 & 0.771 & 0.807 & 0.810 & \textbf{0.815} & 0.813 & 0.799 & 0.801 & 0.811\\
        &5    & 0.773 & 0.822 & 0.803 & 0.836 & \textbf{0.838} & 0.836 & 0.831 & 0.821 & 0.818 & 0.828\\
        &10   & 0.808 & 0.835 & 0.830 & \textbf{0.865} & 0.859 & 0.851 & 0.844 & 0.837 & 0.829 & 0.853\\
		\hline

	\end{tabular}
    }\vspace{-3mm}
\end{table*}


Figure~\ref{fig_accuracy} presents the performance (accuracy) of our approach and the baseline methods on the first two datasets: Mnist and Adult. The f-measure results of entity resolution are presented in Table~\ref{tab_fm}.
Generally, for all the datasets, all the methods converge, except CART, when the label budget is sufficient, e.g. 50\% of the total instances are labeled for training in Mnist and Adult and 5\% in Cora. XG+LTS outperforms all the baselines over all the datasets. The balancing parameter $\alpha$ for the best performance varies, depending on label budgets and datasets. For example, when the label budget is 5\%, XG+LTS with $\alpha = 1$ performs best in Cora and XG+LTS with $\alpha = 0.5$ performs best in DBLP-ACM. When the label budget is relatively small, e.g. less than 1\%, XG+DS achieves a better performance than XG+US in all datasets except for Mnist. When the label budget is larger, e.g. in the range 1\% to 10\%, XG+US performs better than XG+DS. In all cases, CART has the worst performance among all the methods, which is followed by XG+RS.

For the dataset Mnist, both XG+US and XG+LTS obtain better results than the others. The reason why XG+DS does not perform well is due to the large feature space of Mnist. There are in total 784 features in this dataset. Thus, the number of groups is much larger than the number of samples being selected in each iteration, which leads to suboptimal performance. For the dataset Adult, XG+DS performs better than XG+US when the label budget is limited, e.g. less than 0.2\%. However, XG+US achieves better performance when the label budget increases, e.g. more than 1\%. 
For the other datasets, the baselines CART, XG, XG+RS and XG+US have no result when the label budget is small, e.g. 0.01\% in Cora and NCVoter, 0.1\% in DBLP-ACM and DBLP-Scholar. However, both XG+LTS and XG+DS achieve good performance, even when the label budget is small. 


From Figure~\ref{fig_accuracy} and Table~\ref{tab_fm}, we draw the following conclusions: (1) Both uncertainty sampling and diversity sampling contribute to the improvement of the performance. (2) When the label budget is limited, diversity sampling can select informative samples more effectively. However, when the label budget is sufficient, diversity samples are less informative than uncertainty samples.

\medskip
\subsubsection{Cold start problem and class imbalance problem}


As shown in Figure~\ref{fig_accuracy} and Table~\ref{tab_fm}, when the label budget is small, i.e. 0.01\% and less in Cora, 0.5\% and less in NCVoter and DBLP-ACM, and 0.1\% and less in DBLP-Scholar, the methods CART, XG, XG+RS and XG+US have the cold start problem (i.e, the FM values are zero). Compared with these methods, XG+LTS only has the cold start problem in the case that the label budget is 0.1\% in DBLP-ACM. More interestingly, XG+DS does not have the code start problem in all settings of our experiments over all datasets. Since XG+DS is a special case of XG+LTS, this indicates that, when the label budget is small, we can handle the cold start problem by choosing a high value for the parameter $\alpha$.


The four datasets used for entity resolution are highly imbalanced. We can see from Table~\ref{tab_fm} that XG+DS outperforms all the other methods when the label budget is small, while all the baselines have no result. When a dataset is highly imbalanced, samples from the majority class are likely to be selected and samples from the minority class are often ignored, which aggravates the cold start problem. 

\medskip
\subsubsection{Performance under different values of balancing parameter $\alpha$}
Figure~\ref{fig_accuracy} and Table~\ref{tab_fm} show that we have conducted experiments on different values of $\alpha$ (i.e. $\alpha \in \{0,0.5,1,2,5, \infty\}$) over all six datasets. When the value of $\alpha$ increases, the XG+LTS approach biases more on the diversity. When the label budget increases, the XG+LTS approach achieves better performance with a smaller value of $\alpha$. When the budget is low, e.g. less than 0.1\% in Cora dataset, a larger $\alpha$ has a better performance. It indicates that diversity sampling contributes more when the label budget is smaller. On the other hand, when the budget is relatively high, e.g. larger than 5\% in DBLP-ACM and DBLP-Scholar, a smaller $\alpha$ can achieve better performance, and the f-measure results from high $\alpha$ is much smaller, e.g. in DBLP-ACM, the performance of $\alpha = 5$ is about 10\% less than that of $\alpha = 0.5$. It indicates that uncertainty sampling contributes more when the label budget is relatively large. The f-measure results in NCVoter are not distinguishable under various values of $\alpha$ when the label budget is greater than 1\%, since all the f-measure results are similar, i.e. larger than 0.99.

\medskip
\subsubsection{Performance under different sampling distribution methods} 
\begin{figure*} 
		\centering
	\begin{center}
		\includegraphics[width = \textwidth]{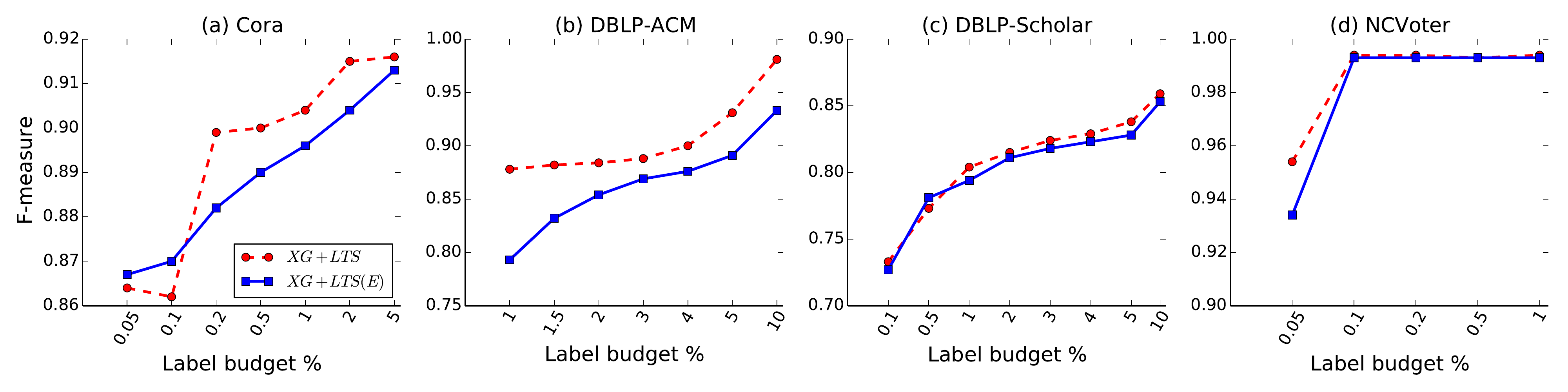}
	\end{center}
\caption{Comparison of f-measure results for the LTS approach under two different sampling distributions} 
	\label{fig_distribution}
\end{figure*}

Now we discuss the experimental results of the LTS approach when using two different sampling distribution methods, i.e. XG+LTS and XG+LTS(E). The experimental results are presented in Figure~\ref{fig_distribution}. We can see that XG+LTS obtains better f-measure results in almost all cases, except for two settings where the label budgets are very small: 0.01\% in Cora and 0.1\% in DBLP-Scholar. This is due to that diversity sampling contributes more in these cases. Therefore, in our LTS approach, we choose eqaul sampling distribution rather than exponentially decreasing sampling distribution.



\medskip
\subsubsection{Comparison of label budgets under the same performance} Table~\ref{tab_sb} presents our experimental results on the four datasets for entity resolution. We set the desired FM value as 0.9 for each dataset, except for the dataset DBLP-Scholar. This is because the dataset DBLP-Scholar is noisy and a classification result with the FM value 0.9 can hardly be achieved. Therefore, we set the desired FM value 0.8 for this dataset. Then we record the amount of label budgets required by each method in order to achieve the desired F-measure values. From Table~\ref{tab_sb}, we can see that, our XG+LTS method ($\alpha=1$) requires the smallest number of samples for each of these datasets, in comparison with the other baseline methods. Especially, for the dataset NCVoter, our XG+LTS approach requires a significantly smaller number of samples for achieving the same performance, in comparison with the baseline methods CART, XG, XG+RS and XG+US. Although XG+DS requires a comparable label budget as our XG+LTS method for the dataset NCVoter, it requires at least a double amount of label budgets for the other three datasets. 

\begin{table}[hb]

\caption{Comparison of label budgets w.r.t. classification results with desired FM values, where XG+LTS has $\alpha = 1$.}
	\centering
	\label{tab_sb}
	\scalebox{1}{
	\begin{tabular}{|c|c|cccc|}
    \hline

    \multicolumn{2}{|c|}{\hspace{0.5cm}{Dataset}\hspace*{0.5cm}} & {Cora} & DBLP-ACM & DBLP-Scholar& {NCVoter}\\
    \hline

    \multicolumn{2}{|l|}{CART} & 5\% & 10\%& 10\%& 3\%      \\
    \multicolumn{2}{|l|}{XG}   & 4\% & 8\% & 2\% & 2\%  \\
    \multicolumn{2}{|l|}{XG + RS} & 5\% & 12\%& 5\% & 2\%  \\
    \multicolumn{2}{|l|}{XG + US} & 2\% & 7\% & 2\% & 7\%  \\
    \multicolumn{2}{|l|}{XG + DS}  & 3\% & 10\% & 2\% & \textbf{0.03\%}  \\
    \multicolumn{2}{|l|}{XG + LTS} & \textbf{0.5\%} & \textbf{4\%} & \textbf{0.9\%} & \textbf{0.03\%}  \\
    \hline
        \multicolumn{2}{|l|}{FM values} & 0.9 & 0.9 & {0.8} & 0.9\\
    \hline
    \end{tabular}
    }
\end{table}

\section{Related Work}

\subsection{Active Learning}

The goal of active learning is to enable a machine learning based model, 
to achieve better performance with relatively fewer but representative training samples, especially when the labels are expensive and very hard to obtain. These samples may be selected from an unlabeled dataset by posing queries and then asking labels from an oracle \cite{settles2010active}. Despite a large number of studies on developing active learning approaches, it is still difficult for a specific task to determine its best-suited one. Thus, meta-learning algorithms have attracted much attention in recent years, driven by the desire to automate the selection process of active learning approaches. For example, Hsu and Lin \cite{hsu2015active} proposed a learning based active learning approach, which allowed a model to adaptively learn from a number of sampling strategies. 

Among various active learning approaches, uncertainty sampling is one of the widely used techniques, which was first proposed by Lewis and Gale \cite{lewis1994sequential}. Normally, uncertainty sampling approaches select samples by measuring their uncertainty, such as probabilistic confidence \cite{culotta2005reducing}, fisher information \cite{settles2010active}, entropy \cite{holub2008entropy} and so on. 
This technique is usually associated with a probabilistic learning model in order to infer labels with the highest probability \cite{kim2006mmr,qian2015relative}. 
A common issue of uncertainty sampling approaches, although computationally efficient and simple to use, is that they do not consider the diversity of data, for example, data with imbalanced class distribution \cite{ertekin2007learning}. 
Furthermore, most of existing uncertainty sampling techniques have the limitation that a sample can be an uncertain sample to one class but a certain sample to another class \cite{jain2009active}.

Diversity sampling is also a useful technique in active learning \cite{brinker2003incorporating, xu2007incorporating}, which aims to select representative samples according to the data distribution. In practice, although uncertain samples are often similar to each other \cite{yang2015multi}, diversity sampling requires samples to be dissimilar in certain features. Thus, samples from different groups or classes are more preferred. 
In our work, we adopt the $l_{2,1}$ norm \cite{jiang2014self} for diversity sampling.

\subsection{Learning based Active Learning}
Two kinds of learning based active learning approaches have been proposed in the literature: One learns to select active learning strategies for a given dataset; The other builds a machine learning model to rank samples for selection.

Hsu and Lin \cite{hsu2015active} proposed \emph{Active Learning by Learning} (ALBL) which relates active learning with multi-armed bandit learner. This approach aims to learn from the performance of a set of active learning strategies so as to decide which is the best. Chu and Lin extended this work by transferring the experience on active learning strategies from one dataset to different datasets \cite{chu2016can}.

The key idea of a recent work called \emph{Learning Active Learning} (LAL) \cite{konyushkova2017learning} is to train a regressor which can predict the generalization error reduction of each unlabelled instance and greedily select one with highest error reduction for labelling. This regressor can be trained as follows: First, given two training sets differing in only one sample, a pair of classifiers is trained, and the corresponding error reduction value of the sample is obtained. Second, the parameters from different pairs of classifiers and the corresponding error reduction values are collected using the Monte Carlo method to train the regressor.
Compared with LAL, our LTS framework captures uncertainty of samples in a learning process w.r.t. a sampling model $G$. More specifically, our LTS framework first predicts samples' probabilities of being mis-classified by a machine learning model $F$, and based on that, a sampling model $G$ is then trained.



There are several other approaches named with ``learning to sample". For example, Li et al. \cite{li2018learning} proposed a generative adversarial network (GAN) based sampling approach which learns to generate synthesized samples by learning likelihood ratios. This approach can also learn to draw samples from an un-normalized distribution via a reference distribution or using Markov Chain Monte Carlo (MCMC). 
Jamshidi et al. \cite{jamshidi2018learning} proposed a transfer learning based approach, which learns the changing of each environment repeatedly for sample selection in configurable software systems. Dovrat et al. \cite{dovrat2019learning} proposed an approach to simplify 3D point clouds by matching them to a fixed size of samples via a learned deep network. 
However, all these approaches do not specifically focus on developing active learning techniques.

\subsection{Boosting Techniques}
A number of \emph{boosting} techniques have been proposed which use a set of weak learners (e.g. decision tree and SVM) to create a single strong learner \cite{kearns1994cryptographic}. Freund developed the first boosting algorithm \cite{freund1995boosting}. Later on, the first adaptive boosting approach, called \emph{AdaBoost}, was proposed \cite{freund1996experiments}, in which the parameters of a model can be self-adjusted based on the actual performance in each iteration, including weights for samples and weights for additive learners. Compared with \emph{AdaBoost}, which favors on dealing with classification tasks, \emph{Gradient Boosting} \cite{friedman2000additive} approaches were proposed to solve both classification and regression problems by reducing the loss of a model in a gradient descent way. 
The state-of-the-art gradient boosting approach is \emph{XGBoost} \cite{chen2016xgboost}. With the use of the sparsity-aware algorithm and the weighted quantile sketch for approximate learning, \emph{XGBoost} can deliver accuracy results efficiently.



\section{Conclusion}

In this paper, we have proposed a novel learning based active learning framework called learning to sample. This framework is composed of a sampling model $G$ and a boosting model $F$. The boosting model is constructed based on a dynamic training set with an increasing number of samples in each iteration. These additional samples are selected iteratively by the sampling model which can learn from the performance of the boosting model through a unified process for two sampling strategies: uncertainty sampling(US) and diversity sampling(DS). The experimental results show that our approach outperforms all the baselines, particularly when the number of samples is relatively small. In addition to this, our framework can handle the cold start problem and the class imbalance problem.

\section*{Acknowledgment}
This work was partially funded by the Australian Research Council (ARC) under Discovery Project DP160101934.

\bibliographystyle{plain}
\bibliography{ICDM19}

\end{document}